%% file: main.tex
\documentclass[10pt,twocolumn,letterpaper]{article}

\usepackage{cvpr}              

\DeclareUnicodeCharacter{2218}{\ensuremath{\circ}}   
\DeclareUnicodeCharacter{00B0}{\ensuremath{^\circ}}  
\usepackage[accsupp]{axessibility}  
\usepackage{booktabs} 
\usepackage{array}    
\usepackage{enumitem} 
\usepackage{multirow} 

\usepackage{float}  
\usepackage[table,xcdraw]{xcolor}
\usepackage{tabularx}
\usepackage{graphicx}
\usepackage{dblfloatfix}  
\usepackage{bm} 
\usepackage{pifont}       

\definecolor{gold}{HTML}{FFD700}    
\definecolor{silver}{HTML}{C0C0C0}  
\definecolor{bronze}{HTML}{CD7F32}  

\newcommand{\capscore}[2]{{\setlength{\fboxsep}{1pt}\colorbox{#1}{\textbf{#2}}}}

\definecolor{cvprblue}{rgb}{0.21,0.49,0.74}
\usepackage[pagebackref,breaklinks,colorlinks,allcolors=cvprblue]{hyperref}
\usepackage{booktabs,makecell}
\def\paperID{6} 
\def\confName{CVPR}
\def\confYear{2026}

\title{MCPDepth: Practical Omnidirectional Depth Estimation from Multiple Cylindrical Panoramas via Stereo Matching}

\author{
Feng Qiao$^{1}$, 
Zhexiao Xiong$^{1}$, 
Xinge Zhu$^{2}$, 
Yuexin Ma$^{3}$, 
Qiumeng He$^{4}$,
Nathan Jacobs$^{1}$ \\
$^{1}$Washington University in St. Louis
$^{2}$The Chinese University of Hong Kong \\
$^{3}$ShanghaiTech University
$^{4}$University of California, Los Angeles
}

\begin{document}
\maketitle
\input{sec/0_abstract}    
\input{sec/1_intro}
\input{sec/2_relatedwork}
\input{sec/3_method}
\input{sec/4_experiments}
\input{sec/5_conclusion}

\section*{Acknowledgments}
We gratefully acknowledge the advanced computational resources provided by Engineering IT and Research Infrastructure Services at Washington University in St. Louis.

{
    \small
    \bibliographystyle{ieeenat_fullname}
    \bibliography{main}
}

\input{sec/X_suppl}
\end{document}

%% file: sec/0_abstract.tex
\begin{abstract}

    Omnidirectional depth estimation presents a significant challenge due to the
    inherent distortions in panoramic images. To address the underexplored impact of projection methods, we introduce Multi-Cylindrical Panoramic Depth Estimation (MCPDepth), a two-stage framework that enhances omnidirectional depth estimation via stereo matching across multiple cylindrical panoramas. MCPDepth initially performs stereo matching using
    cylindrical panoramas, followed by a robust fusion of the resulting depth maps
    from different views. Unlike existing methods that rely on customized
    kernels to address distortions, MCPDepth utilizes standard network components,
    facilitating seamless deployment on embedded devices while delivering
    exceptional performance. To effectively address vertical distortions in cylindrical
    panoramas, MCPDepth incorporates a circular attention module, significantly expanding
    the receptive field beyond traditional convolutions. We provide a
    comprehensive theoretical and experimental analysis of common panoramic
    projections—spherical, cylindrical, and cubic—demonstrating the superior
    efficacy of cylindrical projection. Our method improves the mean absolute error
    (MAE) by 18.8\% on the outdoor dataset Deep360 and by 19.9\% on the real dataset
    3D60. This work offers practical insights for omnidirectional depth estimation and related real-world applications. The code is available at \url{https://github.com/Qjizhi/MCPDepth}.
\end{abstract}

%% file: sec/1_intro.tex
\section{Introduction}
\label{sec:intro}

\begin{figure*}[tb]
  \centering
  \includegraphics[width=1\linewidth]{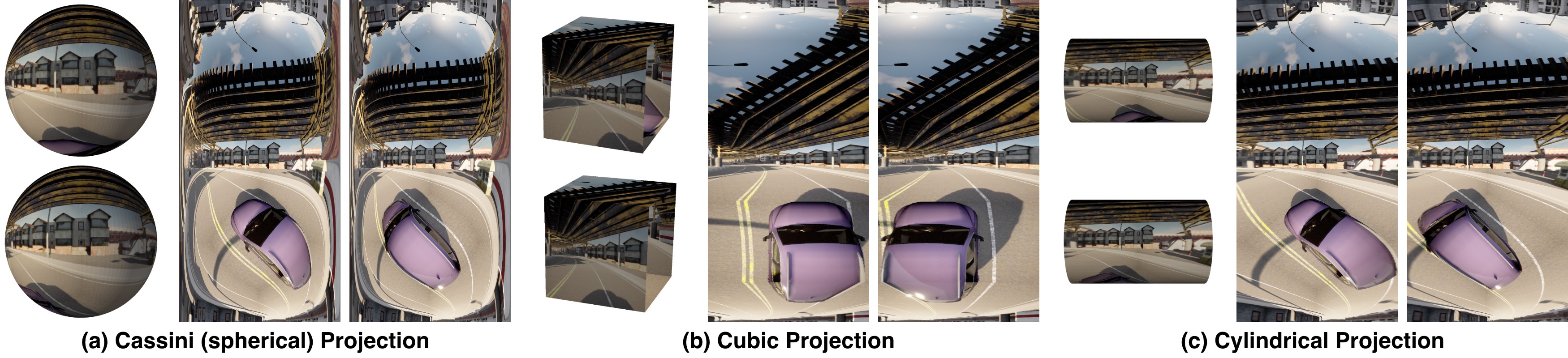}
  \caption{Comparison of stereo images among common panoramic projections.}
  \label{fig:image1}
\end{figure*}

Depth estimation is a pivotal challenge in geometric computer vision, playing a critical
role in 3D scene understanding and robotic perception. Despite substantial
advancements achieved through convolutional neural networks (CNNs) in processing
perspective images, the task of estimating omnidirectional depth remains particularly
challenging due to the severe geometric distortions inherent in panoramic representations.
Recent research has investigated both monocular~\cite{wang2020bifuse,jiang2021unifuse}
and stereo~\cite{wang2020360sd,li2021omnidirectional,li2022mode2} approaches, each
presenting unique advantages and limitations. Methods that apply conventional
CNNs to spherical projections~\cite{wang2020bifuse,jiang2021unifuse,wang2020360sd}
often struggle to effectively manage these distortions, while those that directly
model spherical epipolar geometry~\cite{li2021omnidirectional} encounter significant
computational complexities. Although some strategies have introduced customized
convolutional techniques, such as deformable convolution~\cite{tateno2018distortion},
EquiConvs~\cite{fernandez2020corners}, and spherical convolution~\cite{li2022mode2},
their practical deployment on resource-constrained robotic platforms remains a
formidable challenge~\cite{xu2021energy,su2017learning}. Additionally, the
inherent ambiguities associated with single or dual-view depth estimation
frequently result in unreliable outputs, further complicating the task.

Several works have explored multi-view approaches, such as SweepNet~\cite{won2019sweepnet}
and OmniMVS~\cite{won2019omnimvs}, which use fish-eye cameras to capture a panoramic
field of view (FoV). However, these methods face limitations, including ineffective feature
extraction due to severe radial distortions and ultra-wide FoV when using standard 2D
convolutions, and incomplete depth reconstruction
due to blind spots in fish-eye camera configurations, leading to discontinuities in
the spherical cost volume representation.

Recent research has advanced stereo matching for depth prediction by leveraging
epipolar constraints to reduce the search space and improve accuracy. Notable contributions
include 360SD-Net~\cite{wang2020360sd} and MODE~\cite{li2022mode2}, which have
addressed challenges in complex depth estimation scenarios. MODE, in particular,
introduces a two-stage framework that utilizes Cassini projection~\cite{enwiki:1184209037}
to simplify epipolar geometry, followed by multi-view depth map fusion to
enhance robustness. While these methods achieve state-of-the-art results, they
face computational bottlenecks, especially on resource-constrained devices~\cite{su2017learning},
due to their reliance on spherical convolutions~\cite{coors2018spherenet}. Additionally,
Cassini projection introduces significant distortions, particularly near the poles,
which can degrade depth map quality.

Despite significant advancements in this field, the influence of projection methods
on feature extraction and downstream tasks remains insufficiently explored. As
illustrated in Figure~\ref{fig:image1}, different projections exhibit distinct characteristics,
each influencing the performance of CNNs and downstream tasks. In this work, we
systematically analyze these effects and demonstrate that cylindrical projection
is particularly effective for CNN-based feature extraction.

Drawing from the strengths of stereo matching and two-stage frameworks, we propose MCPDepth, a novel framework that leverages cylindrical projection for stereo matching. Our approach offers three key advantages: it significantly reduces geometric distortion compared to spherical projection, it's compatible with standard 2D convolutions, avoiding computationally intensive spherical convolutions, and it preserves the stereo matching relationship of perspective images, enabling better transfer learning from existing models. Additionally, we introduce a circular attention module that captures long-range dependencies across the full $360{^\circ}$ vertical FoV while mitigating projection-induced distortions. Our contributions can be summarized as follows:

\begin{itemize}
  \item We introduce cylindrical projection for stereo matching and omnidirectional depth estimation and conduct a comprehensive theoretical and experimental analysis comparing
    common projections, highlighting the advantages of cylindrical projection.

  \item We present an innovative circular attention module designed to alleviate
    vertical axis distortions in cylindrical panoramas while significantly enhancing
    the receptive fields of conventional convolutions.

  \item Our method sets new benchmarks on the Deep360 (outdoor) and 3D60 (indoor) datasets.
\end{itemize}

%% file: sec/2_relatedwork.tex
\section{Related Work}
\label{sec:related_work}

\subsection{Deep Learning-based Stereo Matching}

Early methods employed deep neural networks to compute matching costs, such as MCCNN~\cite{zbontar2015computing},
which trains a CNN for initial patch matching costs. Recently, end-to-end neural
networks have dominated stereo matching methods. Works such as~\cite{MIFDB16,
pang2017cascade, liang2018learning, guo2019group, xu2020aanet, li2021revisiting,
tankovich2021hitnet} only use 2D convolutions. Mayer \textit{et al.}~\cite{MIFDB16}
propose the first end-to-end disparity estimation network, DispNet, and its correlation
version, DispNetC. Pang \textit{et al.}~\cite{pang2017cascade} introduce a two-stage
framework named CRL with multi-scale residual learning. GwcNet~\cite{guo2019group}
proposes the group-wise correlation volume to improve the expressiveness of the cost
volume and performance in ambiguous regions.

AANet~\cite{xu2020aanet} adopts a novel aggregation algorithm using sparse points
and multi-scale interaction. Another series of works~\cite{kendall2017end, chang2018pyramid}
use 3D convolutions, which demonstrate great potential in regularizing or filtering
the cost volume. GCNet~\cite{kendall2017end} first implements a 3D encoder-decoder
architecture aimed at regularizing a 4D concatenation volume. PSMNet~\cite{chang2018pyramid}
proposes a stacked hourglass 3D CNN in conjunction with intermediate supervision
to regularize the concatenation volume. Recently, iterative methods~\cite{lipson2021raft,
wang2022itermvs, li2022practical, xu2023iterative} have shown impressive results.
RAFTStereo~\cite{lipson2021raft} proposes to recurrently update the disparity
field using local cost values retrieved from the all-pairs correlations. IGEV-Stereo~\cite{xu2023iterative}
further advances this iterative approach by introducing a geometry encoding volume
to encode non-local geometry and context information. Selective-Stereo~\cite{wang2024selective}
proposes a novel iterative update operator SRU for iterative stereo matching
methods.

In parallel, substantial progress has been made in multi-view stereo (MVS) techniques~\cite{yao2018mvsnet,
yang2020cost, chen2019point, gu2020cascade}, which focus on generating 3D
reconstructions from multiple perspective views, albeit primarily designed for limited-FoV
cameras.

\subsection{Omnidirectional Depth Estimation}

Omnidirectional depth estimation has developed tremendously with neural networks.
Zioulis \textit{et al.}~\cite{zioulis2018omnidepth} present a learning-based
monocular depth estimation method, trained directly on omnidirectional content in
the equirectangular projection (ERP) domain, and later propose CoordNet~\cite{zioulis2019spherical} with a
spherical disparity model. BiFuse~\cite{wang2020bifuse} uses both equirectangular
and cubemap projections for depth estimation. A more effective fusion framework
for ERP and cubemap projection is proposed in Unifuse~\cite{jiang2021unifuse}. Cheng
\textit{et al.}~\cite{cheng2020omnidirectional} introduce a depth sensing system
by combining an OmniCamera with a regular depth sensor. 360SD-Net~\cite{wang2020360sd}
is the first end-to-end trainable network for stereo depth estimation using spherical
panoramas. CSDNet~\cite{li2021omnidirectional} focuses on left-right stereo and
uses Mesh CNNs~\cite{jiang2019spherical} to overcome spherical distortion. SweepNet~\cite{won2019sweepnet}
and OmniMVS~\cite{won2019omnimvs} use multi-view fish-eye images for omnidirectional
depth maps. However, most of them are based on spherical projection and extract spherical
features with regular convolutions, and none of them discuss the properties of
cylindrical projection.

Cheng \textit{et al.}~\cite{cheng2020omnidirectional} propose a spherical feature
transform layer to reduce the difficulty of feature learning. MODE~\cite{li2022mode2}
adopts spherical convolution from Spherenet~\cite{coors2018spherenet}, but the customized
CUDA implementation poses deployment challenges on robotic platforms~\cite{su2017learning}.

Shimamura \textit{et al.}~\cite{shimamura2000construction} employ cylindrical
panoramas for stereo matching, but without CNNs or analysis of cylindrical projection
properties, and 12 perspective images are stitched to obtain the panoramas.

\subsection{Self-Attention Module}
Attention mechanisms were first introduced by~\cite{bahdanau2014neural} for the
encoder-decoder in a neural sequence-to-sequence model to capture token
correspondence between sequences. Self-attention, designed for single contexts, encodes
long-range interactions and has been widely applied in computer vision,
achieving state-of-the-art performance~\cite{vaswani2017attention, parmar2018image,
wang2018non, zhang2019self, huang2019ccnet, Misra_2021_ICCV, Peebles_2023_ICCV, sun2021loftr,
cong2022stcrowd}. Global self-attention in image processing is computationally expensive
due to the need to calculate the relationship between every pixel and every other
pixel, limiting its practical usage across all layers in a full-attention model.
It is shown in~\cite{ramachandran2019stand, hu2019local} that self-attention layers
alone could form a fully attentional model by restricting the receptive field of
self-attention to a local region.

In stereo matching, CREStereo~\cite{li2022practical} first adopts the self-attention
module from LoFTR~\cite{sun2021loftr}. Zhao \textit{et al.}~\cite{zhao2023high}
propose a multi-stage and multi-scale channel-attention transformer to preserve
high-frequency information. GOAT~\cite{liu2024global} uses self-cross attention to
capture more representative and distinguishable features. However, these methods
are not designed for stereo matching in $360{^\circ}$ panoramic images.

More recently, some attention mechanisms~\cite{ling2023panoswin, shen2022panoformer,
yun2023egformer, zhang2025sgformer}
specifically designed for ERP have been proposed. However, these methods cannot
be directly applied to cylindrical projection and face significant deployment
challenges and computational overhead.

%% file: sec/3_method.tex
\section{Method}
Given $m$ $360{^\circ}$ cameras, where $m\ge3$, we have a set of
$n ={m \choose 2}= \frac{m!}{2!(m-2)!}$ pairs of rectified panoramas
$\{(I^{i}_{L}, I^{i}_{R})\}^{n}_{i=1}$ with their intrinsic and extrinsic
parameters. Our objective is to estimate the omnidirectional depth map $d$ for the
left panorama in the first pair $I^{1}_{L}$.

\subsection{Preliminaries: Panorama Projections}
We compare spherical and vertical cylindrical projections for stereo matching. We omit details of the cubic projection, as it follows the same principles as perspective images. Finally, we provide a comprehensive analysis of the advantages and disadvantages of common panoramic projections.

\begin{figure}[tb]
    \centering
    \includegraphics[width=1\linewidth]{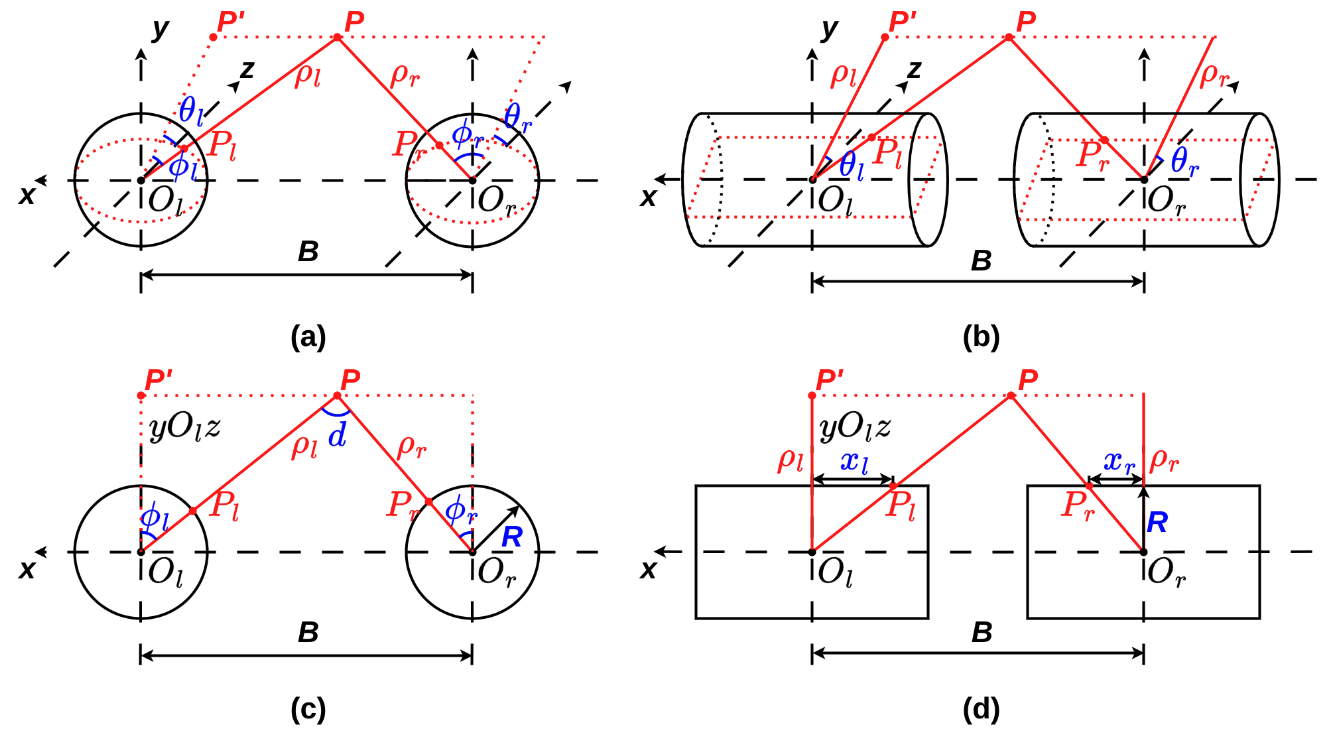}
    \caption{(a) and (b) compare the spherical and cylindrical projections for
    stereo matching and their respective epipolar geometries. (c) and (d) represent
    the schematic drawing of the epipolar plane under spherical and cylindrical projections.}
    \label{fig:image2}
\end{figure}

\begin{table*}[h]
  \centering
  \caption{Comparison of different projection types. $h$ and $v$ represent the horizontal and vertical FoV, respectively.}
  \label{tab:projection_comparison}
  \scriptsize
  \begin{tabular}{@{}l p{7.0cm} p{7.0cm}@{}}
    \toprule
    \textbf{Projection Type} & \textbf{Advantages} & \textbf{Disadvantages} \\
    \midrule
    ERP &
      \makecell[l]{\textbullet~Full coverage: $360^\circ(h)\times180^\circ(v)$ FoV.} &
      \makecell[l]{\textbullet~Non-linear epipolar geometry, complicating stereo matching.} \\
    \cmidrule(lr){2-3}
    Cassini &
      \makecell[l]{%
        \textbullet~Linear epipolar geometry simplifies stereo matching.\\
        \textbullet~Full coverage: $360^\circ(v)\times180^\circ(h)$ FoV.} &
      \makecell[l]{%
        \textbullet~Severe distortion near poles, uneven distortion.\\
        \textbullet~Requires custom kernels for processing.} \\
    \cmidrule(lr){2-3}
    Cubic &
      \makecell[l]{%
        \textbullet~Linear epipolar geometry.\\
        \textbullet~No distortion within individual cube faces.\\
        \textbullet~Compatible with standard convolutional kernels.} &
      \makecell[l]{%
        \textbullet~Limited horizontal FoV: $360^\circ(v)\times90^\circ(h)$.\\
        \textbullet~Discontinuities at cube joints, hindering CNN feature learning.\\
        \textbullet~Requires fusion module across faces.} \\
    \cmidrule(lr){2-3}
    Cylindrical &
      \makecell[l]{%
        \textbullet~Linear epipolar geometry.\\
        \textbullet~No distortion along the horizontal axis.\\
        \textbullet~Uniform distortion along the vertical axis.\\
        \textbullet~Compatible with standard convolutional kernels.} &
      \makecell[l]{%
        \textbullet~Limited horizontal FoV: $360^\circ(v)\times n^\circ(h)$, where $n<180$.\\
        \textbullet~Residual distortion along the vertical axis.} \\
    \bottomrule
  \end{tabular}
\end{table*}

As illustrated in \cref{fig:image2}, both cylindrical and spherical projections
preserve the linear epipolar constraint. In spherical coordinates, $\rho$
represents the Euclidean distance from the origin $O$ to point $P$; $\phi$ is the
angle between line $OP$ and the plane $yOz$; and $\theta$ is the angle between
line $OP '$ and the z-axis, where $P'$ is the projection of $P$ on the plane $yOz$.
In cylindrical coordinates, $\rho$ denotes the Euclidean distance from the x-axis
to point $P$; $\theta$ is the angle between line $OP'$ and the z-axis, where $P'$
is the projection of $P$ on the plane $yOz$. The conversion between spherical, cylindrical,
and Cartesian coordinate systems is illustrated in \cref{eq:projection}.
\begin{equation}
    \begin{aligned}
        \textcolor{black}{\begin{cases}x = \rho \sin(\phi) \\ y = \rho \cos(\phi) \sin(\theta) \\ z = \rho \cos(\phi) \cos(\theta)\end{cases}}
    \end{aligned}
    \begin{aligned}
        \textcolor{black}{\begin{cases}x = x \\ y = \rho \sin(\theta) \\ z = \rho \cos(\theta)\end{cases}}
    \end{aligned}
    \label{eq:projection}
\end{equation}

The spherical and cylindrical panoramas in \cref{fig:image1} (a) and (b) are
generated according to \cref{eq:pixel_coor}, where $u$ and $v$ are pixel coordinates,
$W$ and $H$ are panorama dimensions and $R = H/2\pi$ is the cylinder's radius, which
is the focal length in perspective images. $u$ in the cylindrical panorama is the same as it is in the perspective images.
\begin{equation}
    \begin{aligned}
        {\begin{cases}u = (\phi + \frac{\pi}{2}) \cdot \frac{W}{\pi} \\ v = (\theta + \pi) \cdot \frac{H}{2\pi}\end{cases}}
    \end{aligned}
    \begin{aligned}
        {\begin{cases}u = -\frac{xR}{\rho} + \frac{W}{2} = -\frac{x}{\rho}\cdot\frac{H}{2\pi}+\frac{W}{2} \\ v = (\theta + \pi) \cdot \frac{H}{2\pi}\end{cases}}
    \end{aligned}
    \label{eq:pixel_coor}
\end{equation}
In distortion-free perspective images, an object's actual length and its pixel length
along the horizontal and vertical axes is given by $\Delta u = \frac{f_{x}}{z}\Delta x$
and $\Delta v = \frac{f_{y}}{z}\Delta y$, where $f_{x}$ and $f_{y}$ are the
focal lengths along the $x$ and $y$ axes, and $z$ is the distance along the z-axis.
\cref{eq:delta_pixel} shows these relationships for both spherical and
cylindrical projections.
\begin{equation}
    \begin{aligned}
        {\begin{cases}\Delta u = f_{\phi}\Delta \phi \approx \frac{f_{\phi}}{\rho\cos{\theta}}\Delta X \\ \Delta v = f_{\theta}\Delta \theta \approx \frac{f_{\theta}}{\rho}\Delta Y\end{cases}}
    \end{aligned}
    \begin{aligned}
        {\begin{cases}\Delta u = \frac{f_{X}}{\rho}\Delta X \\ \Delta v = f_{\theta}\Delta \theta \approx \frac{f_{\theta}}{\rho}\Delta Y\end{cases}}
    \end{aligned}
    \label{eq:delta_pixel}
\end{equation}
where $f = R = H/2\pi$. The relation
$\Delta u = f \Delta X / \rho$ and approximation
$\Delta v \approx f \Delta Y / \rho$ for cylindrical projection hold under the
condition that the object is not too large or far from the camera~\cite{plaut20213d}.
This approximation means objects in cylindrical projection appear similar regardless
of their location. This \textbf{shift-invariant property} facilitates efficient learning
by CNNs. In contrast, objects in spherical projection vary with their $\theta$ axis
position, limiting the effectiveness of regular convolutions.

In addition, the disparity in spherical projection is defined as angular
disparity $d$ (\cref{fig:image2} (c)), where $d = \left | \phi_{l}- \phi_{r}\right
|$. This concept has been previously discussed in some works~\cite{li2008binocular,lin2018real,zioulis2019spherical,li2022mode2}.
The relationship between disparity and depth is:
\begin{equation}
    \rho_{l}= B \cdot \frac{\sin(\phi_{r}+\frac{\pi}{2})}{\sin(d)}= B \cdot \left
    [ \frac{\sin(\phi_{l}+\frac{\pi}{2})}{\tan(d)}-\cos(\phi_{l}+\frac{\pi}{2})\right
    ] \label{eq:disp_sph}
\end{equation}
where $B$ denotes the baseline. As shown in \cref{fig:image2} (d), the cylindrical
projection maintains the same disparity-depth relationship as perspective images:
\begin{equation}
    \rho_{l}= \frac{B \cdot f}{\left | x_{l}- x_{r}\right |}
\end{equation}
~\cref{tab:projection_comparison} summarizes the advantages and disadvantages of
different projection types. Cylindrical projection is the most suitable for
stereo matching of panoramas due to the following reasons: (1) Compatibility with
perspective images: Cylindrical panoramas maintain a disparity definition
consistent with perspective images, enabling the direct application of stereo networks
originally designed for perspective images. (2) Reduced distortion: Cylindrical panoramas
only distort vertically, providing better shift invariance, which enhances CNN feature learning. (3) Simplified
Deployment: Spherical panoramas require customized convolutions~\cite{tateno2018distortion,coors2018spherenet,fernandez2020corners}
to extract features. For example, spherical convolutions can't be exported to widely
used ONNX~\cite{enwiki:1186408309} format for deployment, they either need CUDA
Plugins for the TensorRT engine on NVIDIA platforms or customized implementation
on other embedded devices. Cylindrical panoramas only use regular convolutions,
making MCPDepth deployment-friendly.

\subsection{Framework}
\begin{figure*}[th!]
    \centering
    \includegraphics[width=1\linewidth]{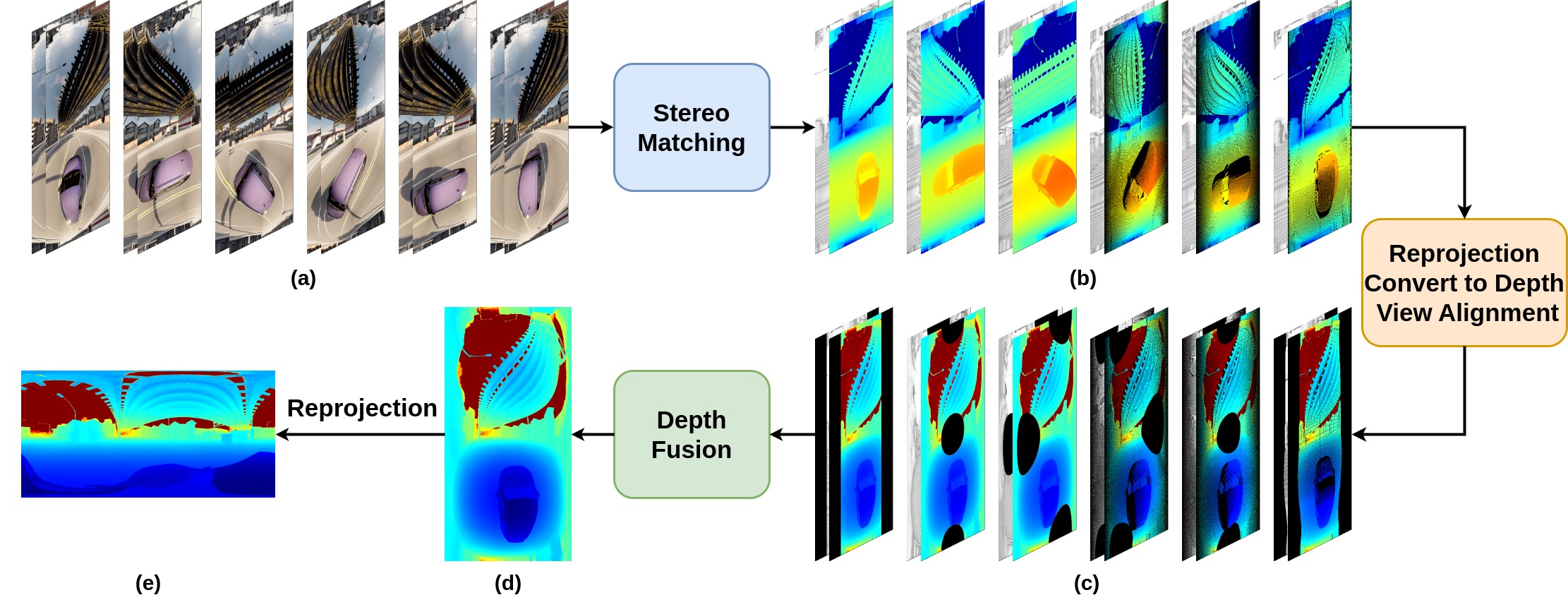}
    \caption{Framework of MCPDepth. (a) represents 6 pairs of cylindrical panoramas,
    (b) shows the disparity and confidence maps, and (c) shows the depth and confidence
    maps. (d) and (e) illustrate the depth map in Cassini and spherical
    projection.}
    \label{fig:image3}
\end{figure*}

The MCPDepth framework, shown in \cref{fig:image3}, includes two stages. In the stereo
matching stage, $n$ pairs of rectified cylindrical panoramas (\cref{fig:image3} (a))
are fed into the stereo matching network. The number of pairs ($n$) and cameras ($m$)
varies on different datasets: $n=6, m=4$ for Deep360, and $n=3, m=3$ for 3D60.
The resulting disparity and confidence maps (\cref{fig:image3} (b)) are
reprojected into the Cassini domain with a $180{^\circ}$ horizontal FoV. The disparity
maps are then converted to depth maps. The depth and confidence maps are aligned
with the view of $I^{1}_{L}$ using extrinsic parameters as shown in
\cref{fig:image3} (c). Black areas indicate invisible and occluded regions.

We use the circular attention module between feature extraction and cost volume
with a structure similar to PSMNet~\cite{chang2018pyramid}. The circular attention
module augments the extracted features to capture features from a $360{^\circ}$ FoV
and overcome vertical-axis distortion. These augmented features are then shifted
and concatenated to build the cost volume. The disparity map is regressed through
the 3D stacked hourglass network. During training, we use the $\ell_1$ loss to train the
network. The confidence maps are used to measure the reliability of the
disparity estimation and are widely used in stereo matching tasks~\cite{poggi2021confidence}.
The confidence map is obtained during inference. Specifically, considering the disparity
is obtained through a probability-weighted sum over all disparity hypotheses, we
compute the corresponding confidence value by taking a sum of probabilities over the
three nearest disparity hypotheses.

We generally follow MODE's depth fusion stage structure. Specifically, multi-view
depth maps, along with their corresponding confidence maps and reference panoramas,
are fed into two separate 2D encoder blocks. The fused depth map is then processed
through a single decoder block, incorporating skip connections between the
encoder and decoder blocks at each scale. The final depth map is generated in the
Cassini domain~\cite{enwiki:1184209037} as shown in ~\cref{fig:image3} (d), a transverse
variant of the ERP commonly used in map projections~\cref{fig:image3}
(e), but it can be readily converted to the ERP domain. More details are available
in the supplementary material.

\subsection{Circular Attention}

\begin{figure*}
    \centering
    \includegraphics[width=1\linewidth]{
        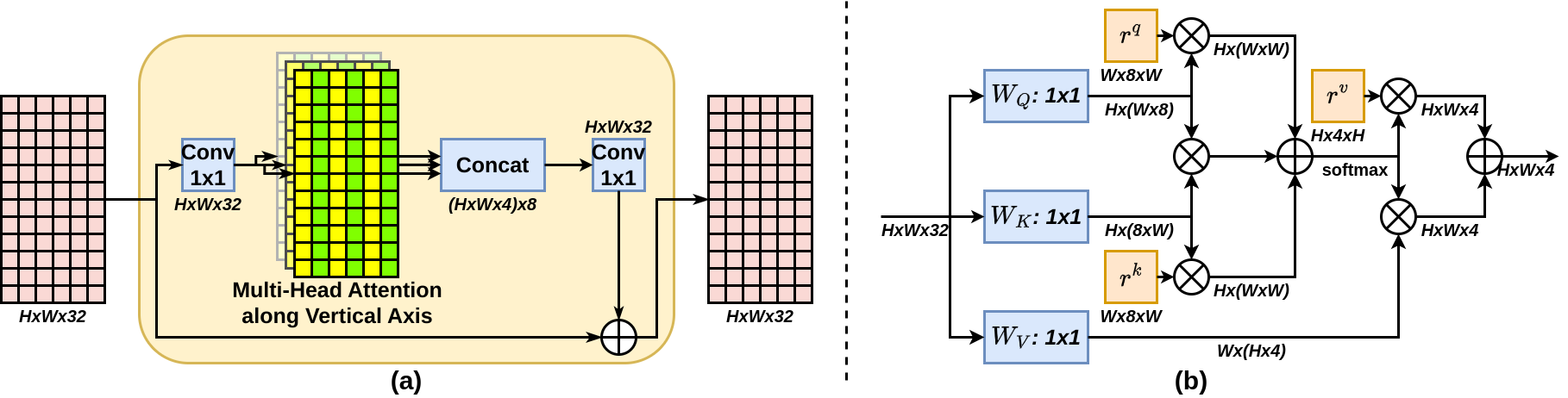
    }
    \caption{(a) displays the circular attention module in the stereo matching
    network. (b) represents our attention applied along the vertical axis.
    $\bigoplus$ denotes element-wise sum. $\bigotimes$ denotes matrix multiplication.
    Blue boxes are $1\times1$ convolution and orange boxes are relative
    positional encoding.}
    \label{fig:image42}
\end{figure*}

To overcome vertical axis distortion and capture the circular $360{^\circ}$ features,
we introduce a circular attention module. Conventional CNNs have limited receptive
fields, which is restrictive for $360{^\circ}$ FoV panoramas. The circular attention
module, placed between feature extraction and cost volume construction is
flexible and can be easily integrated since it maintains the input dimension. Besides,
it only calculates the relations along the vertical axis, conserving more computing
costs compared to global self-attention approaches. \cref{fig:image42} (a)
demonstrates our circular attention module.

In global self-attention, given an input feature map $x \in \mathbb{R}^{h\times w\times
d_{in}}$ with height $h$, width $w$, and channels $d_{in}$. The output
$y_{o}\in \mathbb{R}^{d_{out}}$ at position $o = (i , j)$ can be calculated as:
\begin{equation}
    y_{o}= \sum_{p \in \mathcal{N}}softmax_{p}(q_{o}^{T}k_{p})v_{p}\label{eq:self_attn_1}
\end{equation}
where $\mathcal{N}$ is the whole location lattice, $p=(a,b)$ are all possible
positions. Queries $q_{o}=W_{Q}x_{o}$, keys $k_{o}=W_{K}x_{o}$, and values $v_{o}
=W_{V}x_{o}$ are all linear projections of the input $x_{o}$, where
$\forall o \in \mathcal{N}$. $W_{Q}, W_{K}\in \mathbb{R}^{d_{q}\times d_{in}}$,
and $W_{V}\in \mathbb{R}^{d_{out}\times d_{in}}$ are all learnable weights.

However, global self-attention is extremely resource-consuming and computes $(\mathcal{O}
(h^{2}w^{2}))$. Inspired by ~\cite{ramachandran2019stand,hu2019local}, we
restrict the receptive field of self-attention to a local region and apply only
along the vertical axis. Additionally, global self-attention doesn't contain
positional information, which is proven to be effective in many works~\cite{shaw2018self,ramachandran2019stand,wang2020axial,wu2021rethinking}.
We incorporate positional information in the circular attention module. The output
$y_{o}$ at position $o = (i, j)$ can be calculated as:
\begin{equation}
\begin{split}
    y_{o} &= \sum_{p \in \mathcal{N}_{1\times m}(o)} \text{softmax}_{p} \Big( q_{o}^{T}k_{p} + q_{o}^{T}r_{p-o}^{q} \\
    &\quad + k_{p}^{T}r_{p-o}^{k} \Big) (v_{p}+ r_{p-o}^{v})
\end{split}
\label{eq:self_attn_2}
\end{equation}
%
where $\mathcal{N}_{1\times m}(o)$ is the local $1\times m$ region centered
around location $o=(i, j)$. $r^{q}_{p-o}\in \mathbb{R}^{d_{q}}$ is the learnable
relative positional encoding for queries and the inner product $q^{T}_{o}r^{q}_{p-o}$
measures the compatibility from location $p$ to location $o$. Similarly, the learnable
vectors $r^{k}_{p-o}\in \mathbb{R}^{d_{q}}$ and $r^{v}_{p-o}\in \mathbb{R}^{d_{out}}$
are positional encodings for keys and values. Our circular attention reduces the
computation to $(\mathcal{O}(hwm))$.

For the Deep360 dataset, the feature map size after feature extraction is $h \times
w \times d_{in}= 256\times 128\times 32$. After a $1 \times 1$ convolution is
applied, the feature map is fed into a multi-head attention module, where the
attention mechanism is only applied along the vertical axis. We set span $m = 256$
to ensure it captures all features along the vertical axis. We use 8 heads, each
producing $256 \times 128 \times 4$ outputs. These are concatenated to
$256 \times 128 \times 32$, and after another $1 \times 1$ convolution, the
feature map is added element-wise to the original. \cref{fig:image42} (b) illustrates
how one head of the circular attention module works.

%% file: sec/4_experiments.tex
\section{Experiments}

\subsection{Datasets}
We train and evaluate our framework on Deep360~\cite{li2022mode2} and 3D60~\cite{zioulis2018omnidepth},
which include outdoor and indoor scenes. We evaluate both stereo matching and depth
estimation. For Deep360, four $360{^\circ}$ cameras are arranged horizontally in
a square. Panoramas from all four views are used for evaluation. We use six stereo
pairs for training and testing. For 3D60, three $360{^\circ}$ cameras are arranged
vertically in an equilateral right triangle. Panoramas from two of three views
are used for evaluation. The resolutions are $1024\times512$ and $512\times256$,
respectively.

\subsection{Evaluation Metrics}
Following MODE~\cite{li2022mode2}, we evaluate stereo matching performance using
MAE (mean absolute error), RMSE (root mean square error), Px1,3,5 (percentage of
outliers with pixel error $>$ 1, 3, 5), D1~\cite{pang2017cascade} (percentage of
outliers with pixel error $>$ 3 and $>$ 5\%). We evaluate depth estimation
performance using MAE, RMSE, AbsRel (absolute relative error), SqRel (square
relative error), SILog~\cite{eigen2014depth}(scale-invariant logarithmic error),
$\delta1, 2, 3$~\cite{ladicky2014pulling} (accuracy with threshold that $\max(\tfrac
{\hat{y}}{y^{\star}}, \tfrac{y^{\star}}{\hat{y}})<1.25, 1.25^{2}, 1.25^{3}$).

\subsection{Implementation Details}
We apply nearest-neighbor interpolation for cylindrical/cubic disparity maps
generalization and bilinear interpolation for cylindrical/cubic panoramas
generalization, both derived from spherical inputs.

In the stereo matching stage, cylindrical panoramas have a $360{^\circ}$
vertical FoV and a horizontal FoV of less than $180{^\circ}$. We evaluate the central
part of disparity maps in the Cassini domain with horizontal FoV = $2\arctan(\pi/2) \approx 105{^\circ}$ for both datasets. This FoV yields cylindrical
panorama size equivalent to spherical. In the fusion stage, we evaluate the
entire omnidirectional depth map with a $360{^\circ}$ horizontal and
$180{^\circ}$ vertical FoV.
\subsection{Experimental Results}
\noindent
\textbf{Training on Perspective Images and Testing on Panoramas} The pre-trained
models of PSMNet~\cite{chang2018pyramid} and IGEV-Stereo~\cite{xu2023iterative}
are trained on Scene Flow~\cite{MIFDB16}, which contain only perspective images.
CREStereo~\cite{li2022practical}, trained on mixed datasets, exhibits better
generalization. The performance of stereo matching with different projections on
Deep360 is shown in \cref{tab:pretrained}. Because acquiring panoramic depth ground truth is difficult, these results highlight the value of directly applying perspective-trained models to cylindrical panoramas.

\begin{table}[tb!]
       \centering
       \scriptsize
       \caption{Quantitative results of stereo matching models pre-trained on
       perspective datasets evaluated on the Deep360 test dataset under
       different projections.}
       \begin{tabular}{l|c|c|c|c}
              \toprule Method                                              & Projection  & MAE                     & Px1 (\%)                 & D1 (\%)                  \\
              \midrule \multirow{2}{*}{PSMNet~\cite{chang2018pyramid}}     & Cassini     & 2.7667                  & 42.7912                  & 12.6288                  \\
                                                                           & Cylindrical & \capscore{gold}{2.6118} & \capscore{gold}{34.4403} & \capscore{gold}{10.8204} \\
              \toprule \multirow{2}{*}{IGEV-Stereo~\cite{xu2023iterative}} & Cassini     & 6.5155                  & 61.0948                  & 29.7265                  \\
                                                                           & Cylindrical & \capscore{gold}{4.0194} & \capscore{gold}{53.3429} & \capscore{gold}{22.8117} \\
              \toprule \multirow{2}{*}{CREStereo~\cite{li2022practical}}   & Cassini     & 4.6836                  & 43.5014                  & 18.5130                  \\
                                                                           & Cylindrical & \capscore{gold}{2.1241} & \capscore{gold}{22.6015} & \capscore{gold}{11.2502} \\
              \toprule
       \end{tabular}
       \label{tab:pretrained}
\end{table}

\begin{figure*}[th!]
       \centering
       \includegraphics[width=1\linewidth]{
              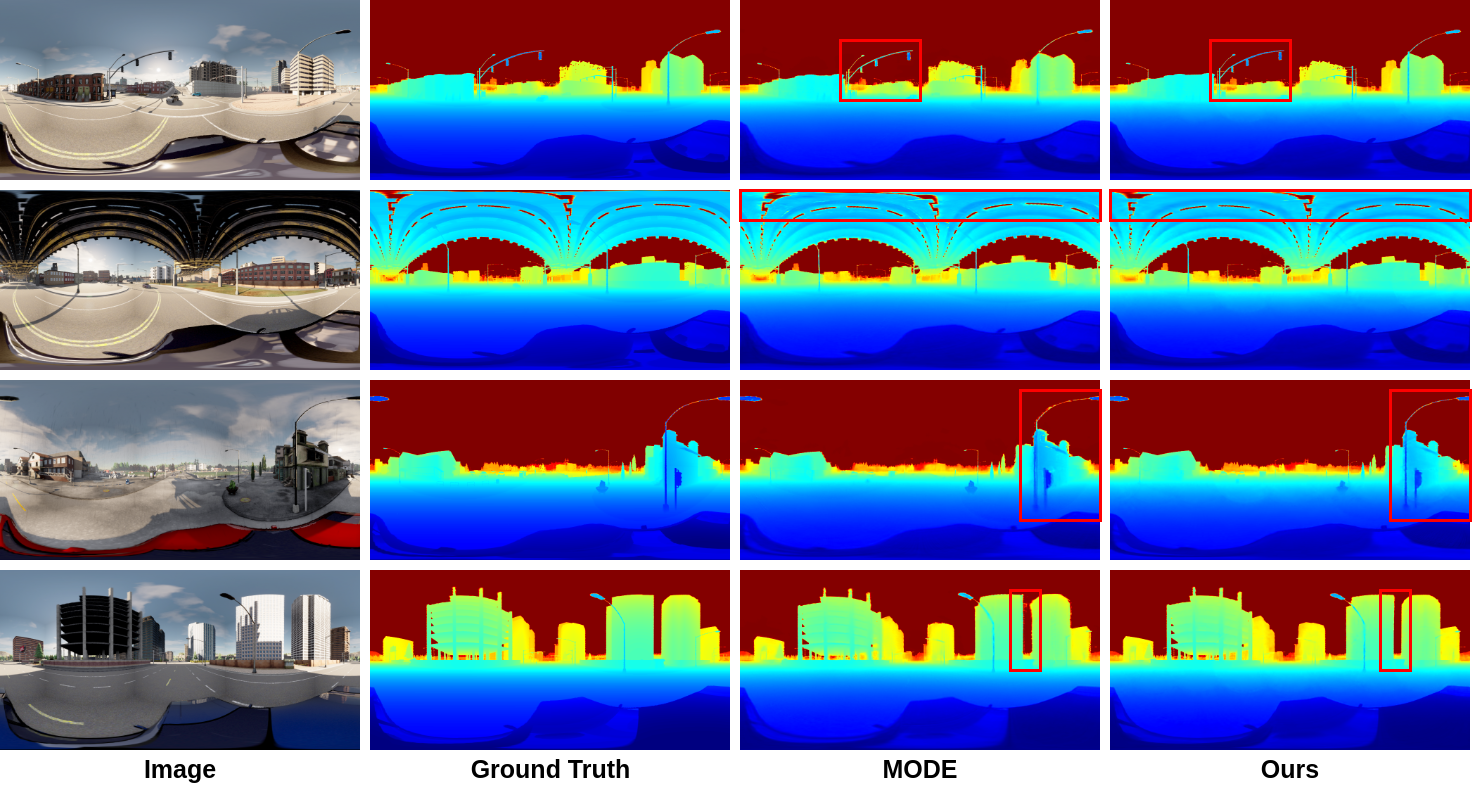
       }
       \caption{Depth estimation results on the Deep360 test dataset. Red boxes highlight regions where MCPDepth preserves finer structural details and sharper foreground boundaries compared to MODE.}
       \label{fig:image5}
\end{figure*}

\noindent
\textbf{Comparisons with State-of-the-Art Methods} We first evaluate our method
against leading stereo matching networks such as PSMNet~\cite{chang2018pyramid},
AANet~\cite{xu2020aanet}, and 360SD-Net~\cite{wang2020360sd}, which is designed for
$360{^\circ}$ stereo. We train these models from scratch and evaluate them on the Deep360 test set following default settings. \cref{tab:state_of_the_art} shows that our method achieves
state-of-the-art performance.

For omnidirectional depth estimation, we compare our method with other multi-view
omnidirectional depth estimation methods including UNiFuse~\cite{jiang2021unifuse},
CSDNet~\cite{li2021omnidirectional}, 360SD-Net~\cite{wang2020360sd}, OmniMVS~\cite{won2019omnimvs},
and MODE~\cite{li2022mode2}. We report the results from MODE.
\cref{tab:state_of_the_art_depth} shows that our method achieves an 18.8\% MAE reduction
on Deep360 and 19.9\% on 3D60 compared to the previous best results, confirming its
effectiveness for diverse real-world panoramas.

\begin{table*}
       [t!]
       \centering
       \scriptsize
       \caption{Quantitative results of stereo matching methods on Deep360 and
       3D60 test datasets. The top three results for each metric are highlighted
       with a \capscore{gold}{first}, \capscore{silver}{second}, and \capscore{bronze}{third}
       background, respectively.}
       \begin{tabular}{l|l|c|c|c|c|c|c|c|c}
              \toprule Dataset                  & Method                         & Projection  & Kernel Type & MAE $\downarrow$          & RMSE $\downarrow$         & Px1 (\%) $\downarrow$     & Px3 (\%) $\downarrow$     & Px5 (\%) $\downarrow$     & D1 (\%) $\downarrow$      \\
              \midrule \multirow{5}{*}{Deep360} & AANet~\cite{xu2020aanet}       & Cassini     & Regular     & 0.3427                    & 1.5703                    & 5.2050                    & 2.1515                    & 1.2847                    & 1.9817                    \\
                                                & 360SD-Net~\cite{wang2020360sd} & Cassini     & Regular     & 0.5262                    & 1.6459                    & 3.8794                    & 1.3389                    & 0.8425                    & 1.2989                    \\
                                                & PSMNet~\cite{chang2018pyramid} & Cassini     & Regular     & \capscore{bronze}{0.2703} & \capscore{bronze}{1.4790} & \capscore{bronze}{3.3556} & \capscore{bronze}{1.1979} & \capscore{bronze}{0.7538} & \capscore{bronze}{1.1708} \\
                                                & MODE~\cite{li2022mode2}        & Cassini     & Spherical   & \capscore{silver}{0.2309} & \capscore{silver}{1.4014} & \capscore{silver}{2.8801} & \capscore{silver}{1.0488} & \capscore{silver}{0.6562} & \capscore{silver}{1.0326} \\
                                                & Ours                           & Cylindrical & Regular     & \capscore{gold}{0.2112}   & \capscore{gold}{1.3903}   & \capscore{gold}{2.5713}   & \capscore{gold}{1.0009}   & \capscore{gold}{0.6376}   & \capscore{gold}{0.9828}   \\
              \midrule \multirow{2}{*}{3D60}    & MODE~\cite{li2022mode2}        & Cassini     & Spherical   & 0.2258                    & 0.5265                    & 2.9441                    & 0.6482                    & 0.2978                    & 0.6478                    \\
                                                & Ours                           & Cylindrical & Regular     & \capscore{gold}{0.1773}   & \capscore{gold}{0.4654}   & \capscore{gold}{2.2298}   & \capscore{gold}{0.5282}   & \capscore{gold}{0.2564}   & \capscore{gold}{0.5279}   \\
              \bottomrule
       \end{tabular}
       \label{tab:state_of_the_art}
\end{table*}

\begin{table*}
       [t!]
       \centering
       \scriptsize
       \caption{Quantitative results of omnidirectional depth estimation methods
       on Deep360 and 3D60 test datasets.}
       \begin{tabular}{l|l|c|c|c|c|c|c|c|c|c}
              \toprule Dataset                  & Method                              & Kernel Type & MAE $\downarrow$          & RMSE $\downarrow$          & AbsRel $\downarrow$       & SqRel $\downarrow$        & SILog $\downarrow$        & $\delta1\% \uparrow$       & $\delta2\% \uparrow$       & $\delta3\% \uparrow$       \\
              \midrule \multirow{6}{*}{Deep360} & OmniMVS~\cite{won2019omnimvs}       & Regular     & 8.8865                    & 59.3043                    & 0.1073                    & 2.9071                    & 0.2434                    & 94.9611                    & 97.5495                    & 98.2851                    \\
                                                & 360SD-Net~\cite{wang2020360sd}      & Regular     & 11.2643                   & 66.5789                    & 0.0609                    & 0.5973                    & 0.2438                    & 94.8594                    & 97.2050                    & 98.1038                    \\
                                                & CSDNet~\cite{li2021omnidirectional} & Spherical   & 6.6548                    & 36.5526                    & 0.1553                    & 1.7898                    & 0.2475                    & 86.0836                    & 95.1589                    & 97.7562                    \\
                                                & UniFuse~\cite{jiang2021unifuse}     & Regular     & \capscore{bronze}{3.9193} & \capscore{bronze}{28.8475} & \capscore{bronze}{0.0546} & \capscore{bronze}{0.3125} & \capscore{bronze}{0.1508} & \capscore{bronze}{96.0269} & \capscore{bronze}{98.2679} & \capscore{bronze}{98.9909} \\
                                                & MODE~\cite{li2022mode2}             & Spherical   & \capscore{silver}{3.2483} & \capscore{silver}{24.9391} & \capscore{silver}{0.0365} & \capscore{gold}{0.0789}   & \capscore{silver}{0.1104} & \capscore{silver}{97.9636} & \capscore{silver}{99.0987} & \capscore{silver}{99.4683} \\
                                                & Ours+Cubic                          & Regular     & 5.0309                    & 36.1907                    & 0.0785                    & 0.4410                    & 0.1781                    & 94.5960                    & 98.1782                    & 98.9406                    \\
                                                & Ours                                & Regular     & \capscore{gold}{2.6384}   & \capscore{gold}{21.6692}   & \capscore{gold}{0.0304}   & \capscore{silver}{0.1153} & \capscore{gold}{0.1033}   & \capscore{gold}{98.2557}   & \capscore{gold}{99.2101}   & \capscore{gold}{99.5227}   \\
              \midrule

\multirow{5}{*}{3D60}   & 360SD-Net~\cite{wang2020360sd}      & Regular     & 0.0762                    & 0.2639                     & 0.0300                    & \capscore{bronze}{0.0117} & 1.4578                    & 97.6751                    & 98.6603                    & 99.0417                    \\
                                                & CSDNet~\cite{li2021omnidirectional} & Spherical   & 0.2067                    & 0.4225                     & 0.0908                    & 0.0241                    & 0.1273                    & 91.9537                    & 98.3936                    & 99.5109                    \\
                                                & UniFuse~\cite{jiang2021unifuse}     & Regular     & \capscore{bronze}{0.1868} & \capscore{bronze}{0.3947}  & \capscore{bronze}{0.0799} & 0.0246                    & \capscore{bronze}{0.1126} & \capscore{bronze}{93.2860} & \capscore{bronze}{98.4839} & \capscore{bronze}{99.4828} \\
                                                & MODE~\cite{li2022mode2}             & Spherical   & \capscore{silver}{0.0713} & \capscore{silver}{0.2631}  & \capscore{silver}{0.0224} & \capscore{silver}{0.0031} & \capscore{silver}{0.0512} & \capscore{silver}{99.1283} & \capscore{silver}{99.7847} & \capscore{silver}{99.9250} \\
                                                & Ours                                & Regular     & \capscore{gold}{0.0571}   & \capscore{gold}{0.1903}    & \capscore{gold}{0.0199}   & \capscore{gold}{0.0027}   & \capscore{gold}{0.0401}   & \capscore{gold}{99.3933}   & \capscore{gold}{99.8506}   & \capscore{gold}{99.9418}   \\
              \bottomrule
       \end{tabular}
       \label{tab:state_of_the_art_depth}
\end{table*}

\cref{fig:image5} illustrates the superior performance on Deep360, effectively
handling severe distortions while preserving finer object details and the edges
between the foreground and background better than MODE.

\begin{figure}[h!]
       \centering
       \includegraphics[width=1\linewidth]{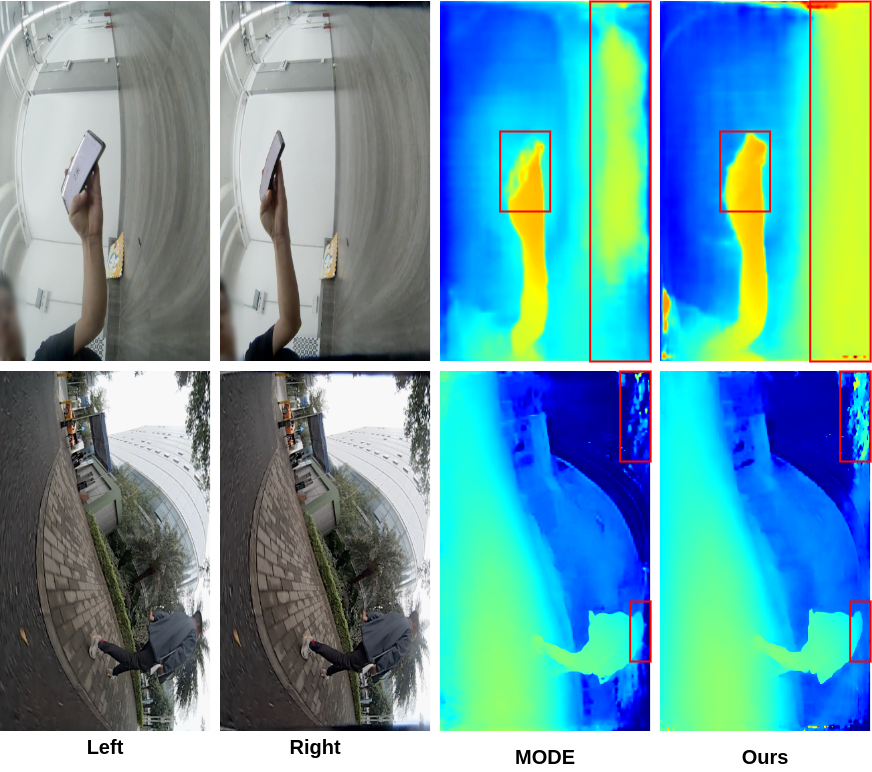}
       \caption{Stereo Matching results in real-world scenarios. Red boxes highlight regions where cylindrical preserves finer structural details and sharper foreground boundaries compared to Cassini.}
       \label{fig:image10}
\end{figure}

\noindent
\textbf{Performance on Real Scenarios} We evaluate our models on real-world
fisheye image pairs. We reproject the fisheye images in Cassini projection,
which have a vertical FoV of $189{^\circ}$ and a horizontal FoV of $120^{\circ}$.
The lens projection is equidistant, and we use OpenCV with checkerboards to
calibrate the camera parameters and the relative pose of the two cameras. ~\cref{fig:image10}
shows the qualitative results of our models compared to MODE. For both indoor and
outdoor scenes, with models trained on 3D60 and Deep360 respectively, MCPDepth demonstrates
noticeable improvements over MODE, particularly in the highly distorted areas. Inference time analysis is provided in the supplementary material, showing that our model achieves comparable speed to MODE while supporting ONNX export for deployment.

\subsection{Ablation Study}


\noindent
\textbf{Panorama Projection} \cref{tab:projection} demonstrates that cylindrical
projection significantly outperforms spherical projection in stereo matching,
even when applying spherical convolutions on spherical panoramas (MODE). Furthermore,
we compare the performance of different projections on depth estimation in
\cref{tab:state_of_the_art_depth}, demonstrating that cylindrical projection is the
most suitable projection for regular convolutions, making it more effective for
panorama stereo matching and depth estimation. These benefits may extend to other
panorama-related vision tasks.

\noindent
\textbf{Circular Attention} \cref{tab:cir_attn} shows that, although designed to
mitigate vertical distortion for cylindrical projection, our circular attention
module consistently improves performance across various panoramic projections
and stereo-matching networks. This lightweight module delivers significant accuracy
gains with minimal additional computation, evaluated in the Cassini domain for
spherical/cylindrical panoramas and the cubic domain for cubic panoramas. For IGEV-Stereo~\cite{xu2023iterative},
applying circular attention to the largest feature map (first scale) yields
substantial performance improvements.

\begin{table}[h]
       \centering
       \scriptsize
       \caption{Ablation study for different projections on the Deep360 test
       dataset. The metrics refer to disparity errors.}
       \begin{tabular}{l|c|c|c|c}
              \toprule Method                                               & Projection  & MAE                     & Px1 (\%)                & D1 (\%)                 \\
              \midrule MODE~\cite{li2022mode2}                              & Cassini     & 0.2309                  & 2.8801                  & 1.0326                  \\
              \midrule

\multirow{2}{*}{PSMNet~\cite{chang2018pyramid}}     & Cassini     & 0.2703                  & 3.3556                  & 1.1708                  \\
                                                                            & Cylindrical & \capscore{gold}{0.2179} & \capscore{gold}{2.6489} & \capscore{gold}{1.0236} \\
              \midrule

\multirow{2}{*}{IGEV-Stereo~\cite{xu2023iterative}} & Cassini     & 0.3905                  & 6.1733                  & 1.8843                  \\
                                                                            & Cylindrical & \capscore{gold}{0.3278} & \capscore{gold}{4.7958} & \capscore{gold}{1.7276} \\
              \bottomrule
       \end{tabular}
       \label{tab:projection}
\end{table}

\begin{table}[h]
       \centering
       \scriptsize
       \caption{Ablation study for circular attention module on the Deep360 test
       dataset. "CA" denotes circular attention. The metrics refer to disparity
       errors.}
       \begin{tabular}{l|c|c|c|c|c}
              \toprule Method                                              & Projection  & CA         & MAE                     & Px1 (\%)                & D1 (\%)                 \\
              \midrule \multirow{2}{*}{MODE~\cite{li2022mode2}}            & Cassini     &            & 0.2309                  & 2.8801                  & 1.0326                  \\
                                                                           & Cassini     & \checkmark & \capscore{gold}{0.2210} & \capscore{gold}{2.7537} & \capscore{gold}{0.9881} \\
              \midrule \multirow{2}{*}{PSMNet~\cite{chang2018pyramid}}     & Cubic       &            & 0.4471                  & 5.0001                  & 1.7623                  \\
                                                                           & Cubic       & \checkmark & \capscore{gold}{0.4196} & \capscore{gold}{4.6699} & \capscore{gold}{1.6464} \\
              \midrule \multirow{2}{*}{PSMNet~\cite{chang2018pyramid}}     & Cylindrical &            & 0.2179                  & 2.6489                  & 1.0236                  \\
                                                                           & Cylindrical & \checkmark & \capscore{gold}{0.2112} & \capscore{gold}{2.5713} & \capscore{gold}{0.9828} \\
              \midrule \multirow{2}{*}{IGEV-Stereo~\cite{xu2023iterative}} & Cylindrical &            & 0.3278                  & 4.7958                  & 1.7276                  \\
                                                                           & Cylindrical & \checkmark & \capscore{gold}{0.2265} & \capscore{gold}{2.9581} & \capscore{gold}{1.1052} \\
              \bottomrule
       \end{tabular}
       \label{tab:cir_attn}
\end{table}

%% file: sec/5_conclusion.tex
\section{Conclusion}

We present MCPDepth, a novel two-stage framework for omnidirectional depth
estimation through stereo matching from multiple cylindrical panoramas. Our theoretical and empirical analyses highlight the distinct advantages of cylindrical projection. Cylindrical projection
maintains the linear epipolar constraint and preserves the definition of
disparity as in perspective images. It effectively reduces distortion, enabling
the application of stereo-matching models trained on perspective images to
cylindrical panoramas. Additionally, cylindrical projection eliminates the need
for customized kernels, simplifying deployment on embedded devices. Our circular
attention module addresses vertical-axis distortions in cylindrical panoramas
and captures $360{^\circ}$ features, and can be extended to other projections.
Experimental results demonstrate that MCPDepth achieves state-of-the-art
performance on both the outdoor dataset Deep360 and the indoor dataset 3D60.

\noindent
\textbf{Limitations} Cylindrical panoramas are limited in their ability to capture
a full $180{^\circ}$ horizontal FoV. As a result, at least 3 cameras are required
to ensure complete coverage. Future work should explore optimizing the
horizontal FoV of cylindrical panoramas to balance performance and computational
resources.

\clearpage

%% file: sec/X_suppl.tex
\clearpage
\setcounter{page}{1}
\maketitlesupplementary

\section{Framework}
\label{sec:framework_supp}

\subsection{Stereo Matching}

\begin{figure}[h]
  \centering
  \includegraphics[width=1\linewidth]{
    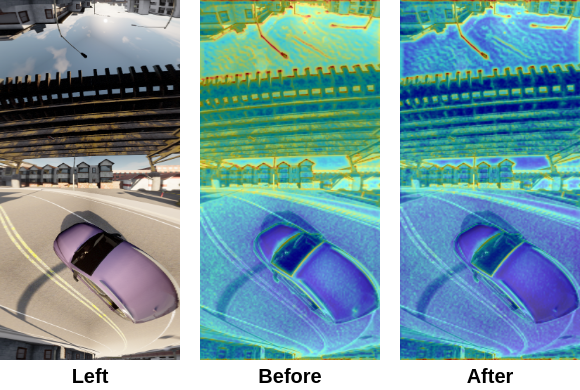
  }
  \caption{Comparison of heatmaps before and after circular attention.}
  \label{fig:attn}
\end{figure}

\begin{figure*}[th!]
  \centering
  \includegraphics[width=1\linewidth]{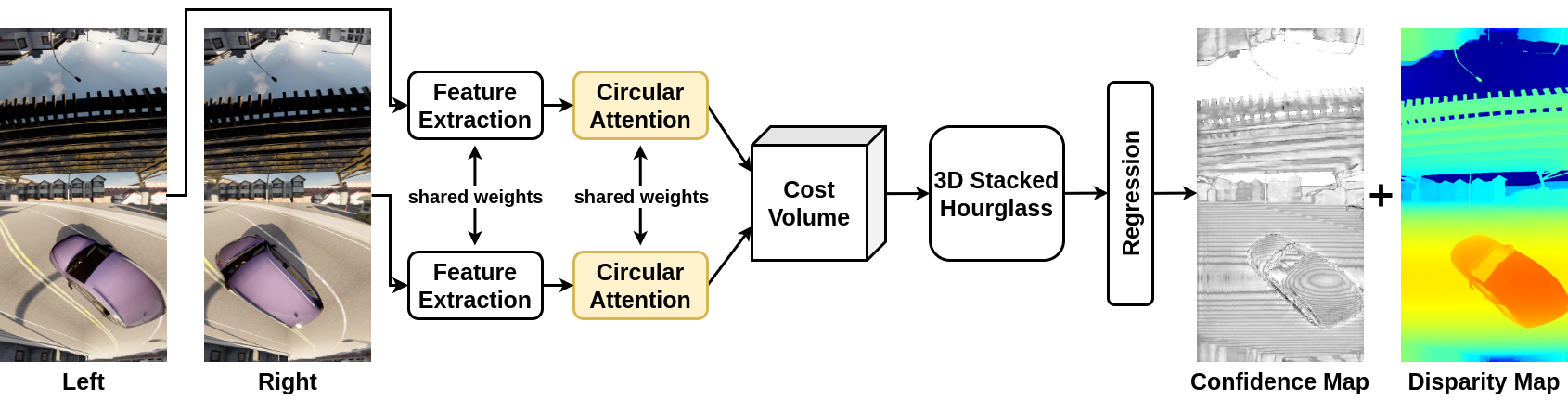}
  \caption{The architecture of the stereo matching network.}
  \label{fig:image41}
\end{figure*}

\cref{fig:image41} illustrates the architecture of the stereo matching network,
incorporating the circular attention module. Meanwhile, \cref{fig:attn} compares
the attention heatmaps before and after applying circular attention. The circular
attention module enhances focus on key regions, particularly the edges of
objects and textured areas where rapid disparity changes occur. In contrast, it
pays less attention to objects' backgrounds and central areas, where disparities
tend to be smooth and consistent. This is because our circular attention module is
capable of capturing the $360{^\circ}$ feature and assigning similar weights to the
same object.

\subsection{Depth Fusion}
\begin{figure*}[tb]
  \centering
  \includegraphics[width=1\linewidth]{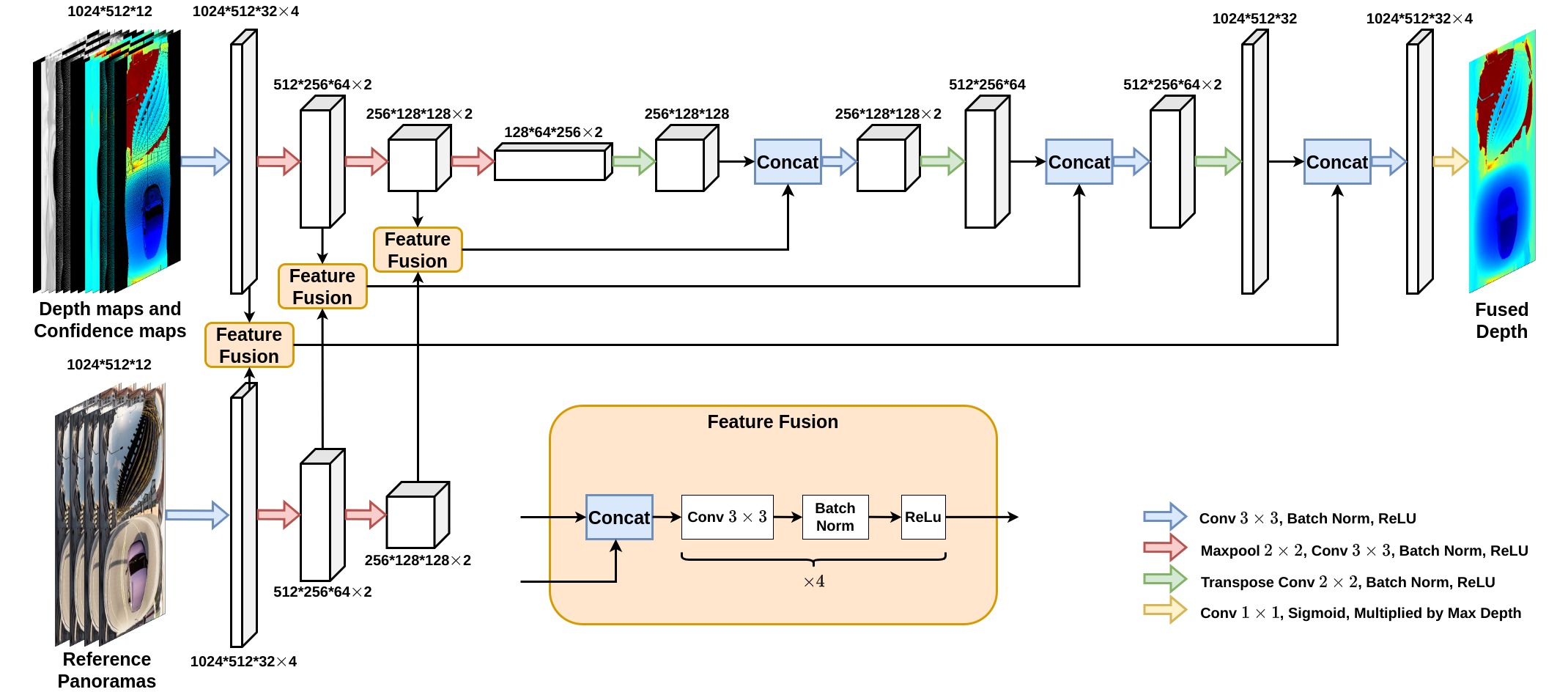}
  \caption{Depth Fusion Architecture.}
  \label{fig:image_depth_fusion}
\end{figure*}

We adopt the depth fusion architecture from MODE~\cite{li2022mode2}, which
integrates depth maps, confidence maps, and reference panoramas for robust depth
estimation. For the Deep360 dataset, 6 depth maps, 6 confidence maps, and 4
panoramas are fed into the fusion model, while for the 3D60 dataset, 3 depth
maps, 3 confidence maps, and 3 panoramas are used. The architecture for Deep360 is
depicted in \cref{fig:image_depth_fusion}.

\section{Training Details}
During the stereo matching stage, we use 2 NVIDIA A40 GPUs to train our models with
a batch size of 4 for the Deep360 dataset. Training takes 158 hours; For 3D60,
we use a single NVIDIA A6000 GPU with batch size 4, which takes 252 hours to train.
The model is trained for 45 epochs with a learning rate of 0.001, followed by a decay
of the learning rate to 0.0001 for an additional 10 epochs. In the depth fusion
stage, we train the network for 150 epochs with a learning rate of 0.0001.

\section{Visualizations}
\begin{figure*}[t!]
  \centering
  \includegraphics[height=21cm]{
    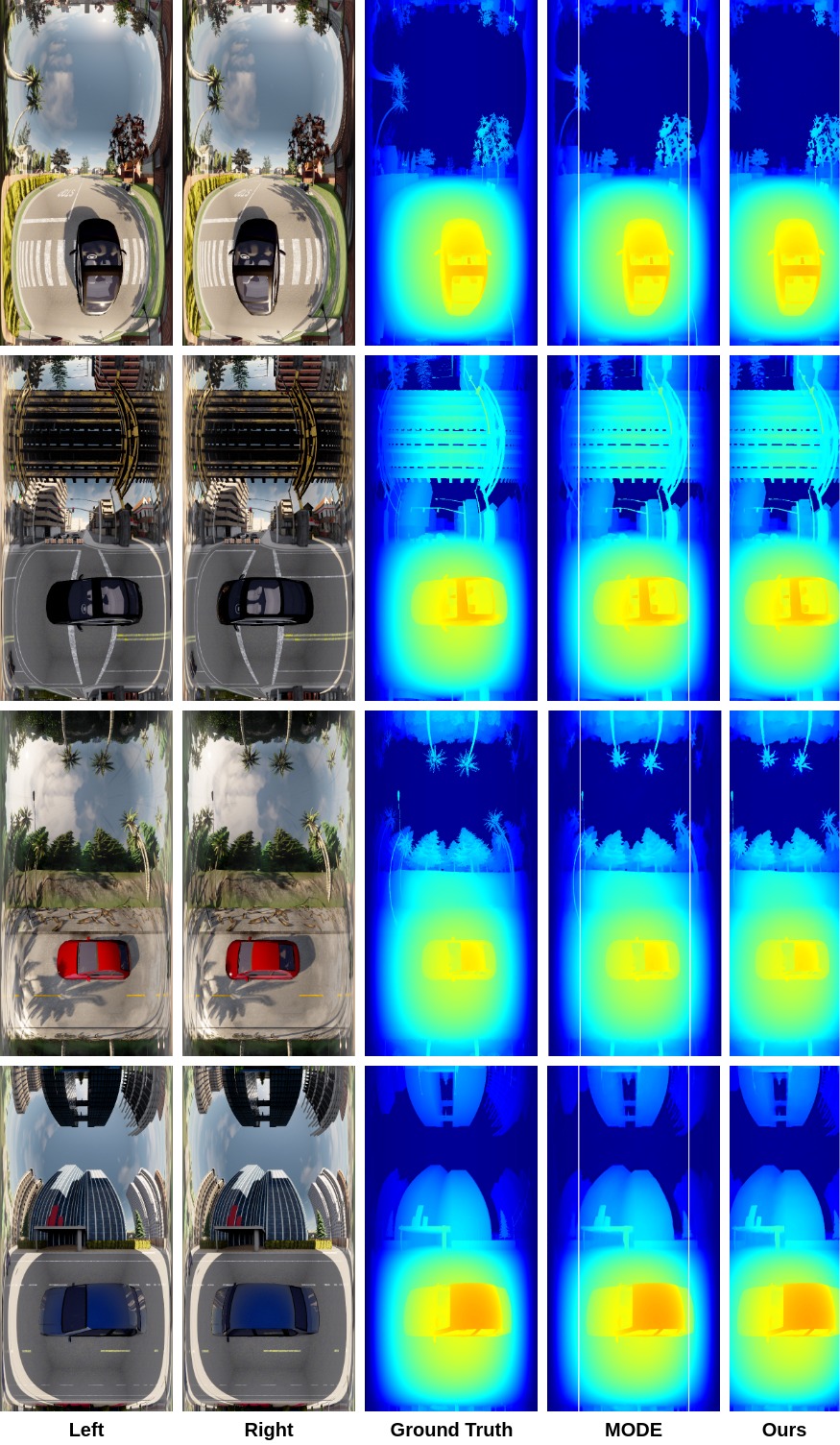
  }
  \caption{Disparity estimation results on the Deep360 test dataset compared to
  MODE.}
  \label{fig:image6}
\end{figure*}

\begin{figure*}[t!]
  \centering
  \includegraphics[height=21cm]{
    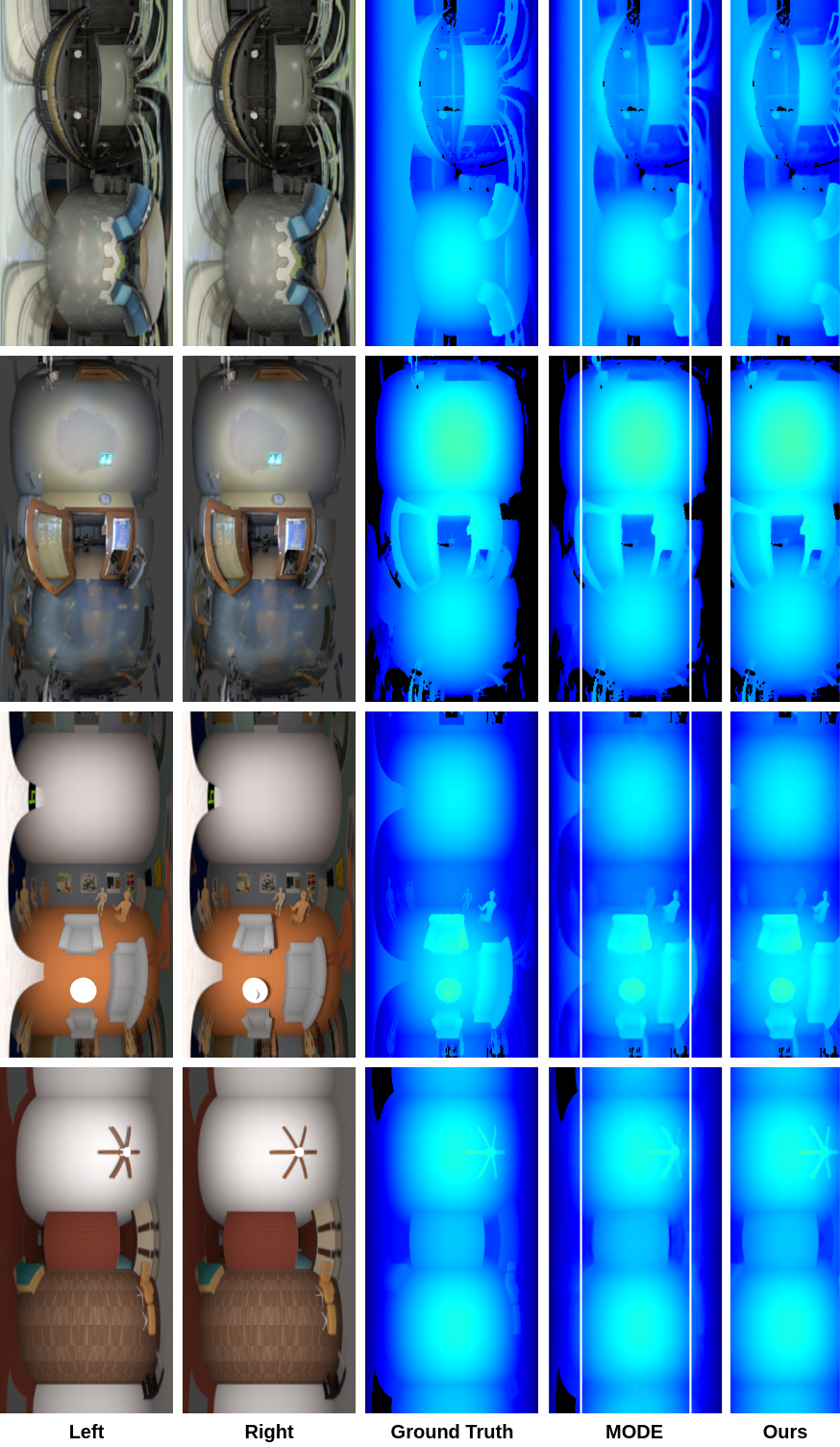
  }
  \caption{Disparity estimation results on the 3D60 test dataset compared to
  MODE.}
  \label{fig:image7}
\end{figure*}

\begin{figure*}[t!]
  \centering
  \includegraphics[width=1\linewidth]{
    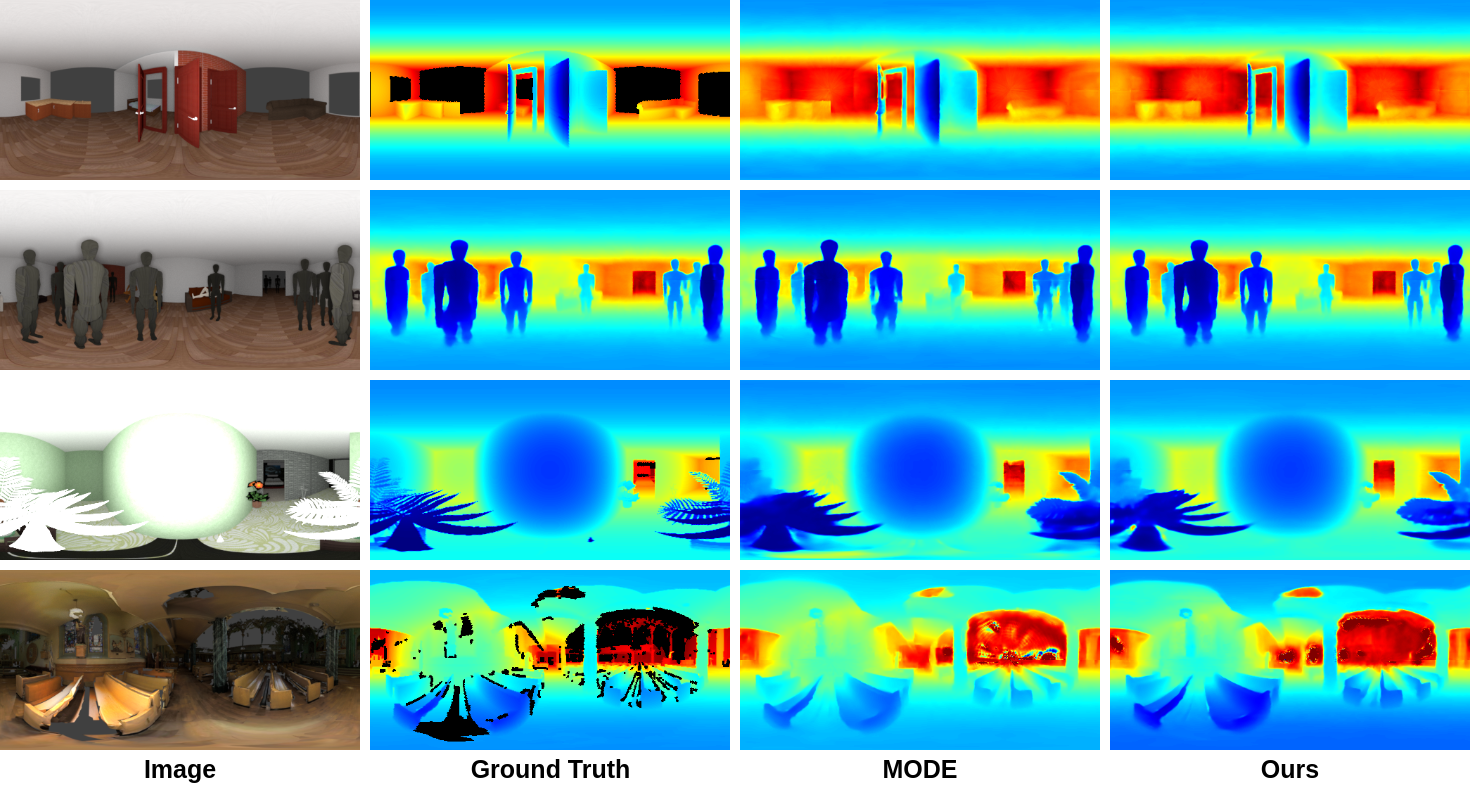
  }
  \caption{Qualitative comparisons of omnidirectional depth estimation methods
  on 3D60 compared with MODE.}
  \label{fig:image8}
\end{figure*}

\cref{fig:image6} and \cref{fig:image7} show the performance of stereo matching on
Deep360 and 3D60 test datasets. Our model demonstrates a strong ability to
distinguish foreground objects from the background, even in regions with significant
distortion. Additionally, \cref{fig:image8} illustrates the depth estimation performance
on the 3D60 test dataset. By leveraging more accurate disparity estimation, our method
surpasses MODE in depth estimation performance, excelling even in areas where
ground truth data is unavailable.

\begin{figure*}[t!]
  \centering
  \includegraphics[width=1\linewidth]{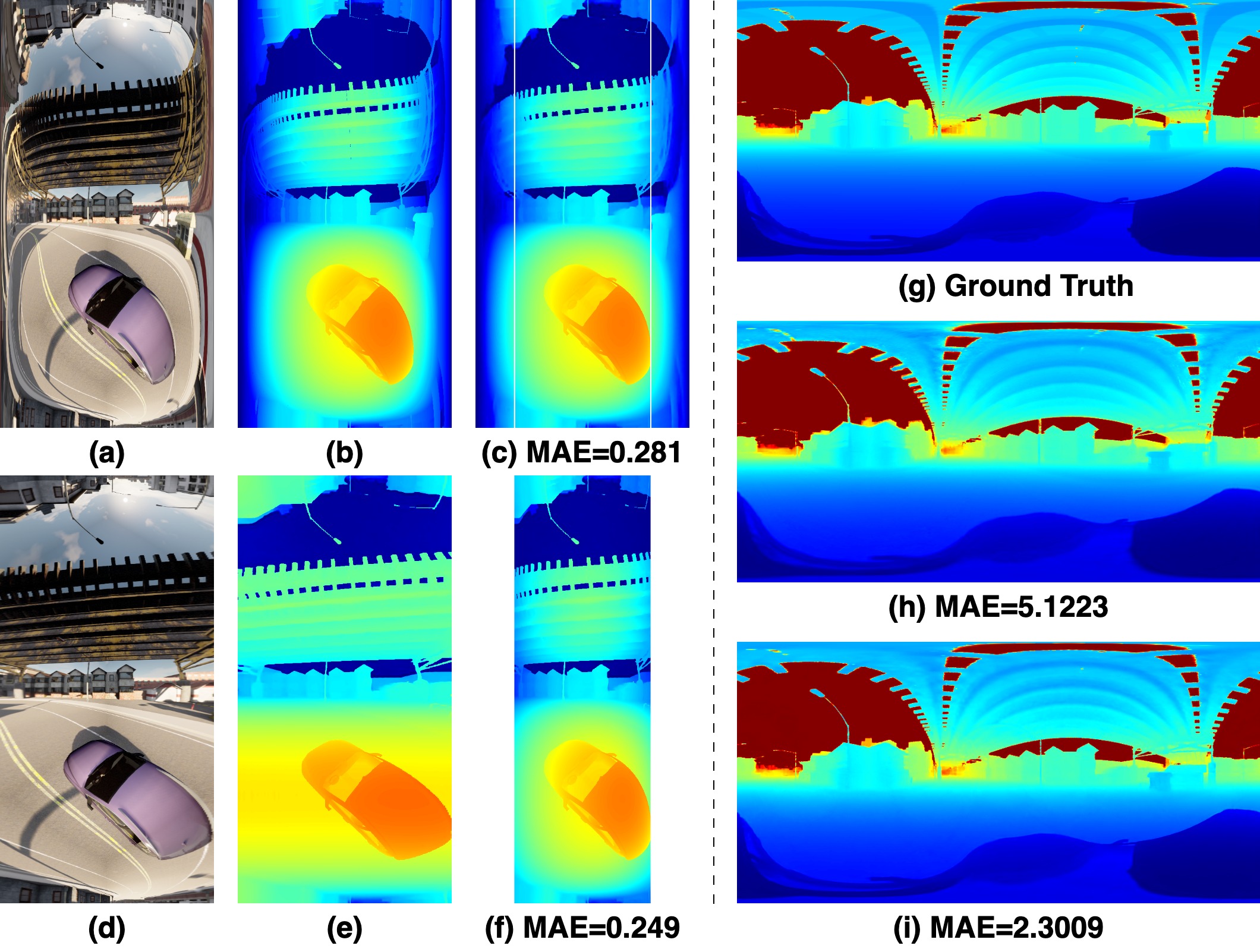}
  \caption{Qualitative comparison with MODE for stereo matching and depth
  estimation. (a) shows the left panorama in Cassini projection, (b) the ground
  truth disparity, and (c) the estimated disparity from MODE. (d) shows the left
  panorama in cylindrical projection, while (e), (f) depict the estimated
  disparity in cylindrical and Cassini projections from ours. (g), (h), (i) are the
  ground truth depth map, the estimation from MODE, and our estimated depth map,
  respectively.}
  \label{fig:performance_comparison}
\end{figure*}

~\cref{fig:performance_comparison} compares the stereo matching and depth estimation
performance of our model with MODE. Our model outperforms MODE in both stereo
matching and depth estimation tasks, achieving higher accuracy and consistency across
the entire dataset.

\section{Attention Mechanism for ERP}
We compare with EGFormer, which is designed for ERP. ~\cref{tab:egformer}
demonstrates that EGFormer hurts cylindrical feature extraction. Our circular attention
designed for cylindrical panorama also improves the extraction of spherical panoramas and cubic panoramas,
demonstrating the generalization of our module.

\begin{table}[h]
  \centering
  \vspace{-2ex}
  \scriptsize
  \caption{Comparison with EGFormer}
  \vspace{-2ex}
  \begin{tabular}{l|c|c|c|c}
      \toprule Projection                   & Method   & MAE             & Px1 (\%)        & D1 (\%)         \\
      \midrule \multirow{3}{*}{Cylindrical} & Baseline & 0.2179          & 2.6489          & 1.0236          \\
                                            & EGFormer & 0.5697          & 10.0792         & 3.7135          \\
                                            & Ours     & \textbf{0.2112} & \textbf{2.5713} & \textbf{0.9828} \\
      \bottomrule
  \end{tabular}
  \label{tab:egformer}
  \vspace{-6ex}
\end{table}

\section{Inference Time of Stereo Matching Models}
The inference time of different stereo matching models on Deep360 and 3D60 datasets
are shown in \cref{tab:infer}. The models are tested in both PyTorch and ONNX
formats on NVIDIA GeForce RTX 3090. The panoramas have dimensions of
$H\times W = 1024\times 512$ on Deep360 and $H\times W = 512\times 256$ on 3D60
for both cylindrical and spherical projections. For Deep360, the maximum
disparity is set to 272 for cylindrical projection and 192 for spherical
projection, while on 3D60, the maximum disparity is set to 256 for all
projections. The circular attention module exhibits inference times of 18.22 ms for
Deep360 and 2.81 ms for 3D60, demonstrating its efficiency across varying
dataset resolutions and disparity ranges.

Our stereo matching model, even without the circular attention module, surpasses
MODE in stereo matching performance. Moreover, it demonstrates a slightly
shorter inference time than MODE when processing the same maximum disparity
range, as shown in \Cref{tab:infer} for the 3D60 dataset. While our model is
marginally slower than MODE in PyTorch format, it offers substantial benefits when
exported to ONNX format. Specifically, our model in ONNX format achieves significantly
faster inference times compared to MODE in PyTorch format when handling the same
maximum disparity range.

\begin{table}[tbh]
  \centering
  \scriptsize
  \caption{Inference time of stereo matching model. "CA" denotes Circular Attention.}

  \begin{tabular}{c|l|c|c|c}
    \toprule Dataset                                          & Methods                 & Projection  & PyTorch (ms) & ONNX (ms) \\
    \midrule \multirow{3}{*}{Deep360~\cite{li2022mode2}}       
                                                              & MODE~\cite{li2022mode2} & Cassini     & 200.62       & -         \\
                                                              & Ours w/o CA             & Cylindrical & 238.61       & 197.63    \\
                                                              & Ours                    & Cylindrical & 266.27       & 226.30    \\
    \midrule \multirow{3}{*}{3D60~\cite{zioulis2018omnidepth}} 
                                                              & MODE~\cite{li2022mode2} & Cassini     & 66.74        & -         \\
                                                              & Ours w/o CA             & Cylindrical & 66.21        & 46.11     \\
                                                              & Ours                    & Cylindrical & 69.12        & 51.24     \\
    \bottomrule
  \end{tabular}
  \label{tab:infer}
\end{table}